# Function Approximation for Reinforcement Learning Controller for Energy from Spread Waves

**Soumyendu Sarkar**[1*] , **Vineet Gundecha**[1] , **Sahand Ghorbanpour**[1] , **Alexander Shmakov**[1] , **Ashwin Ramesh Babu**[1] , **Avisek Naug**[1] , **Alexandre Pichard**[2] and **Mathieu Cocho**[2]

[1]Hewlett Packard Enterprise, USA
[2]Carnegie Clean Energy, Australia
{soumyendu.sarkar, vineet.gundecha, sahand.ghorbanpour, ashmakov, ashwin.ramesh-babu, avisek.naug}@hpe.com, {apichard, mcocho}@carnegiece.com

## Abstract

The industrial multi-generator Wave Energy Converters (WEC) must handle multiple simultaneous waves coming from different directions called spread waves. These complex devices in challenging circumstances need controllers with multiple objectives of energy capture efficiency, reduction of structural stress to limit maintenance, and proactive protection against high waves. The Multi-Agent Reinforcement Learning (MARL) controller trained with Proximal Policy Optimization (PPO) algorithm can handle these complexities. In this paper, we explore different function approximations for the policy and critic networks in modeling the sequential nature of the system dynamics and find that they are key to better performance. We investigated the performance of a fully connected neural network (FCN), LSTM, and Transformer model variants with varying depths and gated residual connections. Our results show that the transformer model of moderate depth with gated residual connections around the multi-head attention, multi-layer perceptron, and the transformer block (STrXL) proposed in this paper is optimal and boosts energy efficiency by an average of 22.1% for these complex spread waves over the existing spring damper (SD) controller. Furthermore, unlike the default SD controller, the transformer controller almost eliminated the mechanical stress from the rotational yaw motion for angled waves.
**Demo:** https://tinyurl.com/yueda3jh

## 1 Introduction

Coastal ocean waves offer a consistent and clean energy source that can help in reducing carbon emissions from power generation. To enhance energy capture and competitiveness, a new design with three generators on three interdependent legs has 2) been developed, utilizing translational and rotational motions. However, this complex design poses challenges for traditional control methods, and the variability of waves in terms of direction, frequency, and height further complicates the process. Additionally, the spinning yaw motion causes high mechanical stress, necessitating measures to prevent costly maintenance in offshore environments.

This work focuses on the CETO 6 Wave Energy Converter (WEC) system, which is currently deployed in Australia and planned for deployment in Europe. Traditional engineering approaches, such as the spring damper, were ineffective in optimizing energy capture efficiency and stress reduction due to the complexities involved. To address these challenges, the paper explores the use of Reinforcement Learning (RL) with a Multi-Agent Reinforcement Learning (MARL) controller and a Proximal Policy Optimization (PPO) training approach. The study evaluates different neural network architectures for the actor and critic and introduces a high-performing STrXL transformer architecture. The goal is to enable optimal control of the generators on the power take-offs (PTOs) to maximize power generation while minimizing mechanical stress and reducing yaw motion, thereby reducing maintenance costs in offshore environments.

The main contribution of this paper can be summarized as follows:

- **First ever implementation of RL controller** for **3-legged** wave energy converter for the **Spread Waves**. No published work has ever addressed this complex real-world problem for an industrial WEC.

- Looks beyond the common RL hyper-parameter tunings and PPO optimizations, into the **different function ap-**

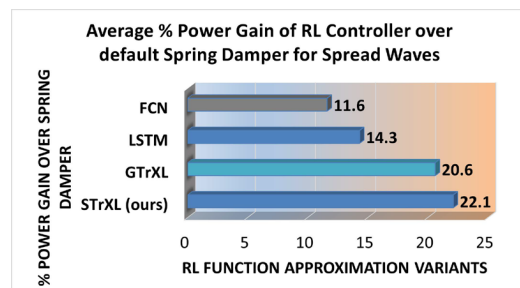

Figure 1: Average Power Gain of RL Controller for different Function Approximations over default Spring Damper Controller

---

*corresponding author

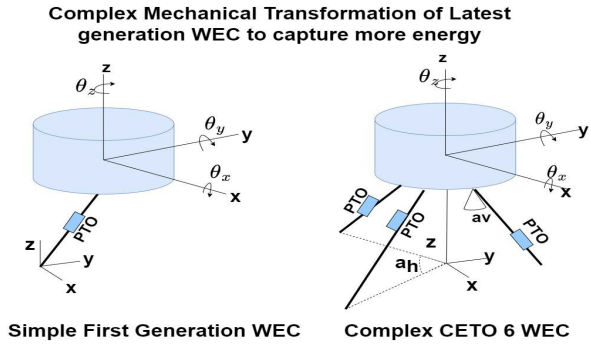

Figure 2: Increase in the structural complexity from 1-tether to the 3-tether WEC to capture more power with 6 degrees of translational and rotational motions. [18]

**proximations** for the policy and critic networks like Fully Connected Neural Network **(FCN), LSTM, and Transformer** variants which can effectively model the sequential nature of system dynamics better.

- Explore and investigate **various gated bypass options** in and around the transformer block for easier trainability and faster convergence with RL which are key bottlenecks in using transformers with RL.
- Propose a transformer block architecture **STrXL** which performs better and trains faster than state-of-the-art GTrXL for this problem- as represented in Figure 1(b) and Figure 10.
- Investigate the effects of various depths of the network for the function approximators.

## 2 Background and Related Work

### 2.1 Wave Energy Converter (WEC)

The CETO 6 WEC is a wave energy converter that uses a cylindrical Buoyant Actuator (BA) to convert the chaotic motions of the ocean into linear motions. The BA is secured to the seabed through three mooring legs, each of which terminates on one of the three power take-offs (PTOs) located within the BA. The PTOs resist the extension of the mooring legs, thereby generating electrical power. The optimal timing of the PTO forces resisting the wave excitation force is key to maximizing WEC performance. Various control strategies like damping control, spring damper control, latching control, and model predictive control exist, attempting to get as close as possible to the optimal force function.

### 2.2 Ocean Waves and Spread Waves

Regular waves also called "monochromatic waves", are simplified waves that comprise of only one period. In other words, each wave is the same as the previous one and the next one and looks like a perfect sinusoidal. Though the ocean at times does exhibit conditions close to regular waves, they mostly have a theoretical interest, as they allow to target the response of a system to a particular frequency excitation

Irregular waves are a superimposition of regular waves of different periods. Which periods are superimposed and how much energy is contained within each period is described by the energy spectrum. Wind waves have a spectrum that peaks at periods below 7 seconds, whereas ground swell is represented by a spectrum that peaks above 7 seconds.

**Spread Waves**: The above descriptions do not mention wave directionality. In the open ocean, waves rarely come from one single direction, but rather from a range of directions as shown in figure 3. In practice, this means that the waves don't look like straight lines propagating toward the observer, but are rather short-crested. The spread factor is representative of how wide a directional range of waves is propagating within. These are called spread waves.

### 2.3 Spring Damper Benchmark Controller (WEC)

The PTO is composed of a mechanical spring and an electrical generator. The damping component is akin to a reactive braking torque against the input shaft, driven by the wave energy source. The captured energy equals the braking mechanical work done by the generator minus losses. The average mechanical power ($\bar{P}_m$) generated by each PTO is the average of the product of the generator force ($F_{gen}$) and the leg extension/retraction velocity ($v_{pto}$).

$$\bar{P}_m = \sum_{i=1}^{3} \overline{F_{gen_i} \times v_{pto_i}} \qquad (1)$$

### 2.4 Related Work

RL has been applied to continuous control tasks for different applications [6] [27], [25], [26], and model-free RL outperforms many model-based non-RL controls. Other unrelated applications of RL include [29], [19], [23], [20], [28], [29], [34], [22], [21], [15], [30], [24], [31]. For wave energy, the bulk of the research on machine learning has focused on wave energy converters (WECs) in academic settings with simplified mechanical designs. A pioneer in this area, [3] used RL in simple one-legged WEC in an academic setting to control the PTO damping and stiffness coefficients for discrete sea states. [1] also applied least-squares policy iteration for resistive control of a nonlinear model of a WEC. [2] also used RL

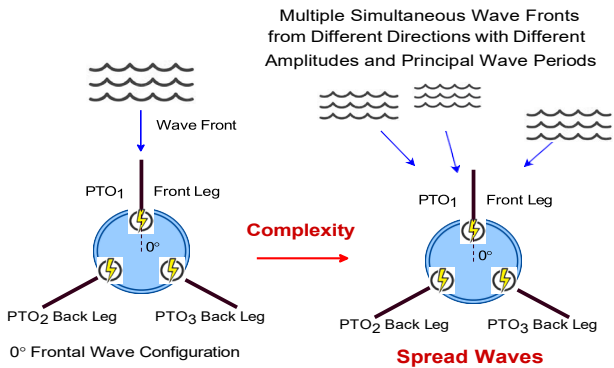

Figure 3: Spread waves significantly increase complexity for WEC with simultaneous waves from different angles based on atmospheric events at different parts of the ocean

to obtain optimal reactive force for a two-body heaving point absorber with one degree of freedom. [38][37] implemented a real-time non-ML control for a simple point-absorbing WEC by designing a wave condition preview-based nonlinear controller, in which the identification of the wave condition was done using an LSTM identifier. [11] also uses LSTM to predict the short-term wave force and a 4$^{th}$ order state-space model to represent the hydrodynamic behavior of the WEC under irregular waves in time-domain with verification using an Autoregressive ANN network. A non-RL technique that utilizes artificial neural networks has also been proposed ([4]) to generate power through WECs. [5] makes use of Deep RL for real-time control of a WEC in continuous action space. However, most of these techniques have been applied to one degree of freedom point absorbers.[7] gives an overview of one-body and two-bodies point absorbers with different controls, and concludes that RL techniques are both effective and viable for controlling the parameters of PTO from an academic perspective. [39] is another example of applying simpler RL DQN algorithm to a point absorber WEC on a small scale. [35] applies Delayed Deterministic Policy Gradient(TD3) RL to Mutriku plant WEC which is of Oscillating Water Column (OWC). In this paper, we apply multi-agent RL to the industrial WEC CETO 6 platform, which has a complex mechanical design of three legs with three generators to capture energy from braking against the motion for all 6 degrees of freedom. **Never before have RL controllers been used for spread waves** under these situations. With waves simultaneously coming from different directions, the asymmetry makes it even more challenging. With the PPO RL controller with fully connected neural network (FCN) RL function approximation, we had limited success. However, in this work, we investigate a variety of ML model architectures for the actor and the critic in the PPO, which can better learn the sequential characteristics of the WEC. Our studies show that these sequential models outperform the FCN by a significant margin. Also, we investigate the different architectural modifications that the transformer model needs to effectively train and converge in a MARL design which has been a major challenge. The skip STrXL trains a lot faster and has a performance that exceeded the state-of-the-art GTrXL. CETO 6 WEC platform is one of the top WEC solutions for the Europe Wave project and can operate at depths where the wave energy is more persistent and takes wave energy closer to financial viability.

## 3 Reinforcement Learning for Wave Energy Converters

### 3.1 RL Algorithm Exploration

After exploring different RL algorithms like Deep Q-Learning (DQN) [14], Soft Actor-Critic (SAC) [8], and Asynchronous Advantage Actor-Critic (A3C) [13], we limited our focus to Proximal Policy Optimization (PPO) ([32]) for this study. We found through evaluation that PPO outperformed other models for this WEC controller. DQN has problems with continuous action space for WEC, and both SAC and A3C had training stability issues and performed poorly. We fine-tuned the PPO implementation to achieve stable convergence of MARL. But these refinements were not enough for the best performance without refining the underlying function approximation of the actor and critic.

### 3.2 Multi-Agent RL Choice

In WECs, the generators mounted on the individual legs tend to generate different amounts of energy based on the orientation of the mechanical structure and variation of wave directionality. Simpler, single-agent RL with multiple actions failed to control the WEC effectively and resulted in poor performance. To accommodate the heterogeneity, we needed to use separate agents (MARL) to control the individual generators on each of the legs. This enabled convergence to a better policy. The architecture is shown in Figure 4.

### 3.3 RL State and Action Design

We validated the inclusion of states with ablation studies of total rewards. Figure 5 shows the attributes included in the **state**. The continuous **action space ($a_i$)** for the individual RL agent indexed "$i$" is defined by the reactive force ($F_{gen_i}$) for the controlled generator.

$$a_i = F_{gen_i} \quad (2)$$

### 3.4 Reward: Cooperation vs. Competition for Agents

Even though each RL agent controls the generator on an individual leg, it needs to consider power contributions from all the generators in its reward for overall WEC optimization. An individual RL agent can be cooperative or competitive based on the effectiveness of the trade-off. The reward function for power is represented in Equation 3.

$$P_{reward} = P_{own} + \eta \cdot P_{others} \quad (3)$$

Here, $P_{own}$ is the generator's power being controlled, and $P_{other}$ is power from other generators, and $\eta$ = team coefficient. A negative "$\eta$" implies adversarial contributions of power from the other legs in the reward for the competitive, while a positive value implies cooperation. With a combined Bayesian hyperparameter search for the multi-agent, we achieved the best performance with $\eta$ of 0.8 for the agents for the back legs and -0.6 for the agent for the front leg.

### 3.5 Design for Trust

The spinning yaw motion of the voluminous buoy of WEC causes the tether connections to wear out and has potential maintenance implications. The yaw is high for angled wavefronts w.r.t. the axis of symmetry of WEC. The incentive to reduce yaw is added to the **total reward** of the RL agents as a weighted addition of the power:

$$Reward = a.P_{reward} + (1 - a).yaw \quad (4)$$

where $a$ is a tunable yaw penalty hyper-parameter (lower the stronger), and $P_{reward}$ is defined in Equation 4. This led to significant improvements in yaw reduction compared to the currently deployed spring damper controller.

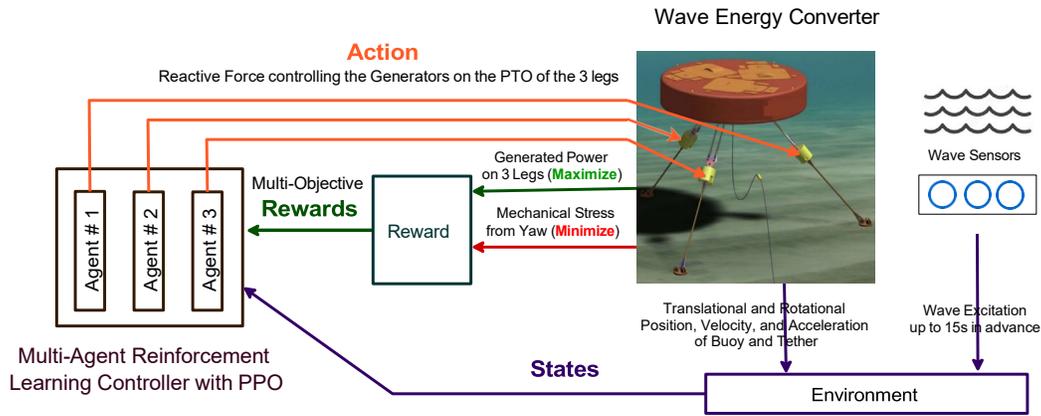

Figure 4: Architecture of Multi-Agent RL controlling the WEC

### 3.6 Exploration

Exploration is key to ensuring that the policy does not converge to a local optimum. PPO explores by sampling from the action distribution. However, this is dependent on the standard deviation predicted by the policy. We tried two exploration techniques - parameter space noise [17] and state entropy maximization with random encoders [33]. According to our studies, parameter space resulted in better performance than just sampling from the action distribution.

## 4 Function Approximation for RL

The periodic nature of ocean waves and the inertia and spring-type response of the WEC require a system model which can represent and process sequential information. So the RL policy and critic function approximators need to have an architecture to support this, unlike the default feed-forward network. LSTMs can represent long-term information using a recurrent architecture. The Transformers have tremendous success in the ability to process a sequence by explicitly ensuring interaction between the elements of the sequence with the attention mechanism. However, unlike LSTMs, they are limited by the limited sequence length. The Tr-XL architecture solves this problem by keeping a memory of hidden states corresponding to previous sequences. But unlike supervised learning tasks like Natural Language Processing and Computer Vision, the difficulty in training remains a key challenge of using transformers in RL.

| | |
|---|---|
| position | position of the buoy with velocity and acceleration for the translational and rotational motion |
| yaw | rotational yaw motion to monitor stress |
| tether | extension and velocity of tether |
| wave | wave elevation and rate of change for present and 10s ahead in time from sensors |

Figure 5: RL States

### 4.1 Function Approximation for Policy and Critic Network

Refinements beyond the RL agent algorithm and hyper-parameters depend on function approximators (FA), such as the most suitable deep neural networks, to leverage the exploration-exploitation trade-off and hence efficiency at the core of RL. Function approximation blends statistical estimation issues with dynamic optimization issues, resulting in the need to balance the bias-variance tradeoffs that arise in statistical estimation with the exploration-exploitation trade-offs that are inherent in RL. These changes are inspired by the objective to combine long-term behavior from past observations and future observations of wave states from sensors placed further into the ocean, into the representation of the current state with the predictive power of the short-term memory transformer architectures.

### 4.2 Function Approximators Explored for WEC

We investigated the WEC controller performance and speed of convergence of fully connected neural networks (FCN), LSTMs, and Transformers of varying depths ([10; 36; 16; 12]). Transformers with the attributes like multi-head attention, temporal convolution network, and large contextual horizon with relative position encoding, are ideally suited for PPO function approximation for WEC.

### 4.3 Exploring Transformer Architectures for Function Approximation

We also explored the effect of variation of gated bypass for Transformer FA on the stability and speed of training in the RL setting, as the standard transformer models are too unstable to train and learn outside supervised learning. With identity map reordering in TrXL-I the layer normalization is placed on only the input stream of the submodules ([16]). This enables an identity map from the transformer's input at the first layer to the transformer's output after the last layer, unlike canonical transformers, where there are a series of layer normalization operations that non-linearly transform the state encoding. The state encoding is passed un-transformed to the policy and value heads, enabling the agent to learn a Markovian policy at the start of training. In our case, the reactive behaviors need to be learned before memory-based ones can be effectively utilized. We also found that GRU style multiplicative interactions with powerful gating mechanisms in place of the residual connections within the transformer block helped stabilize learning and improved performance.

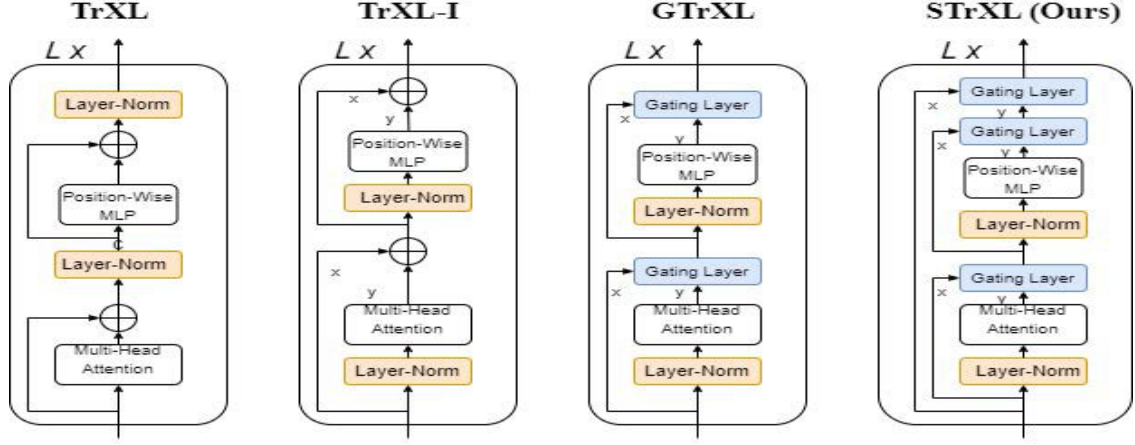

Figure 6: Variations of Transformer blocks to facilitate convergence. The overall architecture has L such blocks. **TrXL:** A traditional transformer with multi-head attention and layer normalization, **TrXL-I:** The layer norm is included with the input stream, **GTrXL:** Includes a gating layer to include the residual connection around the attention and MLP blocks ([16]), **STrXL:** Residual connections with gating layer aroud the transformer block.

### 4.4 Skip Transformer-XL (STrXL) Architecture

Unlike supervised learning tasks ([36]), using traditional transformers as function approximators for a RL task is extremely difficult to optimize, as established by [16]. Inspired by residual network architecture ([9]) and expanding earlier work of [16], we propose a variant of the transformer "Skip Transformer-XL" or STrXL, that enables faster convergence during training specifically in an RL setting represented in Figure 6. An additional bypass connection with a gating layer around the transformer block helps accelerate training convergence when compared to the previous methods like GTrXL, as presented in Figure 10 showing the training progression of these models. The GRU gating adds to the non-linearity. The equations below represent the Gated Recurrent Unit type gating.

The STrXL block can be represented with the following equations:
The Gated Recurrent Unit (GRU) type gating:

$$r = \sigma(W_r^{(l)} y + U_r^{(l)} x)$$
$$z = \sigma(W_z^{(l)} y + U_z^{(l)} x - b_g^l)$$
$$\hat{h} = \tanh(W_g^{(l)} y + U_g^{(l)} (r \odot x))$$
$$g^{(l)}(x, y) = (1 - z) \odot x + z \odot \hat{h}$$

The input to the transformer block is $E^{(l-1)}$ The output of the transformer block is $E^{(l)}$, where l is the layer index. The multi-head attention (MHA) block gated output:

$$\bar{Y}^{(l)} = MHA(LayerNorm(M^{(l-1)}, E^{(l-1)})),$$
$$Y^{(l)} = g_1^{(l)}(E^{(l-1)}, ReLU(\bar{Y}^{(l)}))$$

The MLP block gated output:

$$\bar{E}^{(l)} = f^{(l)}(LayerNorm(Y^{(l)})),$$
$$E^{(l)} = g_2^{(l)}(Y^{(l)}, ReLU(\bar{E}^{(l)}))$$

The STrXL gated output:

$$\bar{\bar{E}}^{(l)} = g_3^{(l)}(E^{(l-1)}, ReLU(\bar{E}^{(l)}))$$

## 5 Experiments

The CETO 6 wave energy converter (WEC) platform simulator was used to accurately model the mechanical structure, the mechanical response, the electro-mechanical conversion efficiency with losses for generator and motor modes, and the fluid dynamical elements of the wave excitation.

Wave data such as the distribution of principal time periods, height, and spectrum were collected from Albany, and Garden Island in Western Australia, Armintza in Spain Biscay Marine Energy Platform, and Wave Hub on the north coast of Cornwall in the United Kingdom. The wave generator model used in simulation uses a well-established ocean wave spectrum like Jonswap, which accurately models the heterogeneous components in ocean waves, letting the simulator sample the waves for training and evaluation. For evaluation, we used 1000 episodes for each principal wave period and height, where each episode covers 2000 sec of continuous wave data in steps of 0.2 sec for RL loop and 0.05 sec (4x) for simulation response. Each training run has roughly 50 million steps for convergence, with 2000 training runs required for hyper-parameter optimization and model search with early stops. For regular operation, we show results of a median wave height of 2m for the entire wave frequency spectrum spanning time periods of 6s to 16s.

Spread Waves: RL % Gain of Energy Capture over default (SD controller)
% Gain for Wave Height = 2m

| Wave Time Period(s) | 6 | 7 | 8 | 9 | 10 | 11 | 12 | 13 | 14 | 15 | 16 | Avg |
|---|---|---|---|---|---|---|---|---|---|---|---|---|
| FCN | 15.2 | 15.4 | 12.0 | 11.7 | 12.2 | 10.2 | 13.5 | 8.4 | 9.2 | 10.1 | 9.4 | 11.6 |
| LSTM | 18.2 | 19.2 | 15.2 | 14.2 | 15.2 | 13.2 | 11 | 11 | 12.5 | 15.1 | 12.1 | 14.3 |
| GTrXL | 22.2 | 24.1 | 25.4 | 23.9 | 19.3 | 14.9 | 23.2 | 15.1 | 17.4 | 19.9 | 21 | 20.6 |
| **STrXL (ours)** | 23.1 | 25.2 | 24.2 | 25.2 | 21.4 | 22.3 | 25.4 | 17.2 | 20.2 | 20.5 | 18.2 | **22.1** |

Table 1: Spread waves: Energy Capture Gain by the RL controller over Spring Damper controller for different PPO function approximators

RL % Gain of Energy Capture over default Spring Damper (SD controller)
% Gain for Wave Height = 2m, and Wave Angle = 0 degrees

| Wave Time Period(s) | 6 | 7 | 8 | 9 | 10 | 11 | 12 | 13 | 14 | 15 | 16 | Avg |
|---|---|---|---|---|---|---|---|---|---|---|---|---|
| FCN | 38.4 | 35.4 | 23.0 | 19.7 | 15.1 | 14.1 | 13.5 | 11.9 | 12.9 | 11.7 | 11.4 | 18.8 |
| LSTM | 41.3 | 35.5 | 27.8 | 24.1 | 18.6 | 15.4 | 15.9 | 17 | 17.7 | 15.9 | 15.3 | 22.2 |
| GTrXL | 40.2 | 36.1 | 28.2 | 23.9 | 19.3 | 14.9 | 23.2 | 17.9 | 18.9 | 18.3.2 | 15.8 | 23.8 |
| **STrXL (ours)** | 40.1 | 38.9 | 32.2 | 25.2 | 21.4 | 22.3 | 24.1 | 18.5 | 19.0 | 18.1 | 17.1 | **25.2** |

% Gain for Wave Height = 2m, and Wave Angle = 30 degrees

| Wave Time Period(s) | 6 | 7 | 8 | 9 | 10 | 11 | 12 | 13 | 14 | 15 | 16 | Avg |
|---|---|---|---|---|---|---|---|---|---|---|---|---|
| FCN | 33.4 | 32.9 | 20.1 | 9.6 | 5.3 | 7.6 | 10 | 14.6 | 15.8 | 16.1 | 12.3 | 16.2 |
| LSTM | 34.6 | 33.3 | 26.7 | 20.3 | 14.5 | 14.3 | 16.3 | 20.8 | 17.9 | 20.1 | 18.7 | 21.6 |
| GTrXL | 39.2 | 35.2 | 27.6 | 21.8 | 17.1 | 17.8 | 21.3 | 29.4 | 36.9 | 30.6 | 31.9 | 28.1 |
| **STrXL (ours)** | 39.7 | 34.7 | 28.7 | 22.7 | 17.4 | 18.1 | 22.6 | 31.5 | 38.7 | 31.2 | 31.3 | **28.8** |

Table 2: Unidirectional waves: Energy Capture Gain by the RL controller over spring damper for different PPO function approximators

## 6 Results

The power generated by the baseline spring damper controller with resonant spring constant and damping constant is used as a reference for evaluation to estimate RL controllers' gain of energy capture as a percentage improvement. We use both the complex spread waves and unidirectional waves of different time periods and heights as shown in Figure 3. We used the same seed for sampling waves for episodes between RL and SD for evaluation.

### 6.1 Energy Capture Gains and Variations with PPO Function Approximation

Table 1 shows that for spread waves, the MARL with STrXL performs on an average 22.1% better than the baseline spring damper (SD) controller. In contrast, the LSTM performs 14.3% better and FCN performs 11.6% better on average for the entire range of principal wave periods 6s to 16s. Even though the STrXL is 1.5% better than GTrXL overall, it trains much faster with high stability. Table 3 shows the variance of the energy capture for the spread waves with the RL controller.

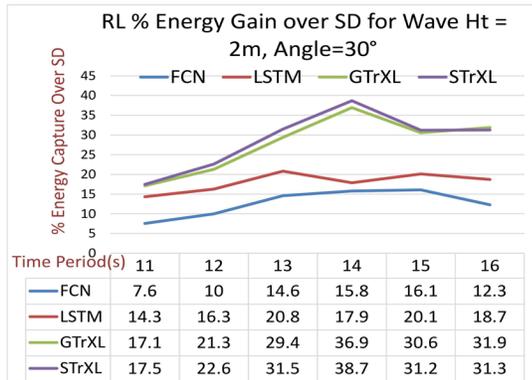

Figure 7: % Increase of Energy Capture over SD controller for 30° waves for Height=2m with different Function Approximations.

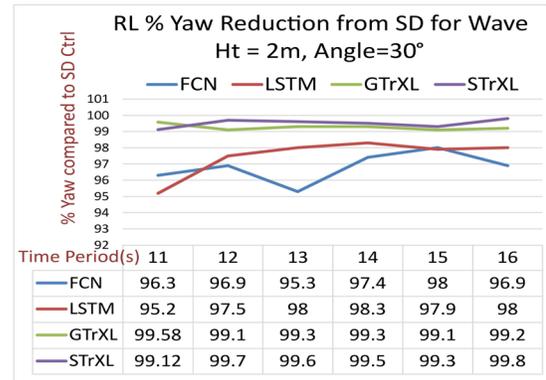

Figure 8: % Yaw Reduction over SD controller for 30° waves for Height=2m with different Function Approximations.

| WTPs | 11 | 12 | 13 | 14 | 15 | 16 | Avg |
|---|---|---|---|---|---|---|---|
| Std. Dev. | 2.3 | 2.8 | 3.0 | 2.9 | 2.5 | 2.4 | 2.65 |

Table 3: Variance of WEC Energy capture for Spread Waves with RL controller

| WTPs | 11 | 12 | 13 | 14 | 15 | 16 | Avg |
|---|---|---|---|---|---|---|---|
| % yaw ↓ | 98.8 | 98.8 | 98.9 | 99.1 | 98.9 | 98.6 | 98.6 |

Table 4: % reduction of Yaw by RL over SD for ht=7m, angle = 30°

Table 2 show that for 0° uni-directional frontal waves, the MARL with STrXL performs on an average 25.2% better than the baseline spring damper (SD) controller, while the LSTM performs 22.2% better and FCN performs 18.8% better on an average for the entire range of wave time periods 6s to 16s. For angled waves of 30°, the MARL with STrXL (28.8%) performs much better than LSTM (21.6%) on average.

## 6.2 Mechanical Stress and Yaw Minimization under Normal and Survival Conditions

Figure 3 shows the results of yaw reduction with the reward shaping for aggressive Yaw minimization for $a$=0.2, where 80% weightage is given to penalty for rms yaw, and 20% weightage is given to energy capture maximization. For the entire range of wave time periods of 11s to 16s, where yaw is a significant problem of default SD controller, the yaw is reduced by more than 99%, significantly reducing mechanical stress with huge maintenance savings. The power generation increased with aggressive yaw control as an indirect effect. Even under dangerous conditions with an extreme wave height of 7m. Table 4 shows that the PPO with STrXL function approximation reduces the yaw by over 98.6% for the critical range of wave time periods from 11s to 16s, where the baseline SD controller faces high yaw.

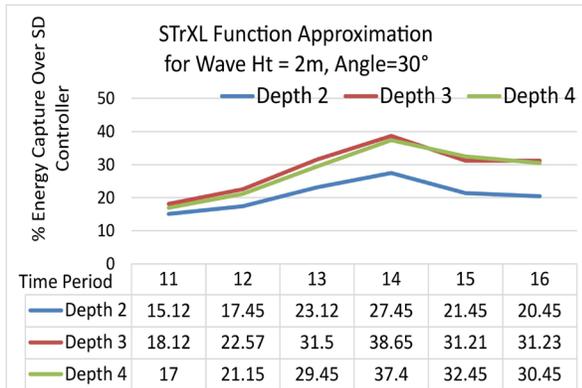

Figure 9: % Increase of Energy Capture over SD controller for 30° waves for Height=2m with STrXL with a variation of a number of layers or depth.

## 6.3 Variation With a Depth of Models for Function Approximators

Figure 9 shows that in the best-performing STrXL transformer model, the performance peaks at a depth of 3. For LSTM, both the depths of 2 and 3 yielded similar results. However, for FCN a depth of 2 has an overall better performance, even though the FCN performs worse than STrXL, GTrXL, and LSTM.

## 6.4 Training Convergence for Different FA Models

Figure 10 compares the average RL environment steps required for convergence for different Transformer function approximators across several time periods. STrXL converged the fastest compared to GTrXL and TrXL-I.

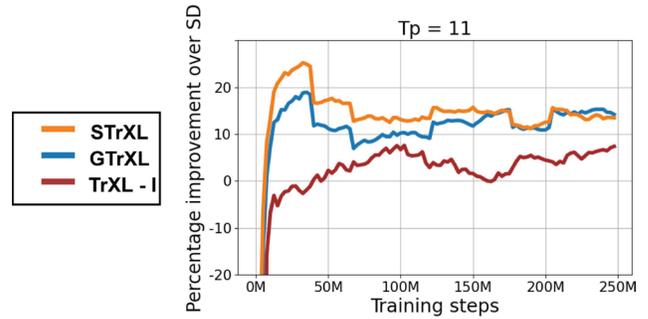

Figure 10: RL Training progression for ht=2m angle=30° for STrXL, GTrXL, and TrXL-I function approximators.

## 7 Conclusions

The proposed MARL controller with STrXL transformer function approximation yields 22.1% gain over the baseline Spring Damper controller (SD) on an average for the entire spectrum of spread waves, boosting energy production with revenue implications. The MARL also helped reduce mechanical stress, which impacts maintenance and operating costs, and actively mitigated the adverse effects of high waves. This is the first published paper where an RL controller can successfully control a 3-legged WEC for spread waves while beating the baseline spring damper controller.

We found that robust RL function approximation sequence models of suitable architectures and depths are key to achieving higher performance for complex real-life use cases like WEC, and RL refinements alone cannot do that. Also, the biggest challenge with the highest-performing transformers is training convergence, and GRU-gated bypass inside and around the transformer block help solve this problem. The proposed novel STxRL architecture trains faster and performs better than the state-of-the-art GTrXL. The STrXL model and PPO function approximation exploration and analysis for MARL will help others to stabilize training convergence for complex RL control systems.